\documentclass[letterpaper, 10 pt, journal, twoside]{IEEEtran}
\IEEEoverridecommandlockouts                           
\usepackage{cite}
\usepackage{subfigure}
\usepackage{graphicx}
\usepackage{textcomp}
\usepackage{xcolor}
\usepackage{multirow}
\usepackage{enumitem}
\usepackage{booktabs}
\usepackage{tikz}
\usepackage{amsmath,amssymb,mathtools,amsfonts}
\usepackage{algorithm,algorithmic}
\usetikzlibrary{arrows.meta,positioning,fit,calc,shapes.geometric,shapes.misc}
\usepackage{makecell}
\usepackage[T1]{fontenc}
\usepackage[cmintegrals]{newtxmath}
\usepackage{bm}
\usepackage{soul}
\usepackage{array}
\usepackage[protrusion=true,expansion=true]{microtype}
\setlist[itemize]{leftmargin=*,nosep}

\let\oldthebibliography\thebibliography
\renewcommand{\thebibliography}[1]{%
  \oldthebibliography{#1}%
  \setlength{\itemsep}{0pt}%
  \setlength{\parsep}{0pt}%
  \setlength{\parskip}{0pt}%
}

\title{\LARGE \bf
Physics-Informed Sliding-Window Particle Filtering for Tactile-Only In-Hand 6-DoF Object Pose Refinement
}
\author{Lingjun Shao$^{1}$, Ying Zhang$^{2}$, Xiangfei Li$^{1}$,\emph{~Member,~IEEE}, Xiangyang Li$^{2}$, Huan Zhao$^{1}$,\emph{~Member,~IEEE},\\ Zhenyu Wang$^{1}$ and Han Ding$^{1,3}$,\emph{~Senior Member,~IEEE}
    \thanks{Manuscript received: March 23, 2026; Revised: June 19, 2026; Accepted: August 4, 2026.}
    \thanks{This paper was recommended for publication by Editor Lorenzo Jamone upon evaluation of the Associate Editor and Reviewers' comments.
    This work was supported by the Fundamental and Interdisciplinary Disciplines Breakthrough Plan of the Ministry of Education of China under Grant JYB2025XDXM208, and the National Science Foundation of China under Grants 52575574, 52188102 and U24A20130, and the Guangdong Power Grid Co., Ltd. Project under Grant 0300002025030103JX00026. (Corresponding author: Xiangfei Li. Email: {\tt\footnotesize lixiangfei@hust.edu.cn.} Lingjun Shao and Ying Zhang contributed equally to this work.)}% 
    \thanks{$^{1}$Lingjun Shao, Xiangfei Li, Huan Zhao, Zhenyu Wang and Han Ding are with the State Key Laboratory of Intelligent Manufacturing Equipment and Technology, School of Mechanical Science and Engineering, Huazhong University of Science and Technology, Wuhan 430074, China. }%  
    \thanks{$^{2}$Ying Zhang and Xiangyang Li are with the Electric Power Research Institute, Guangdong Power Grid Co Ltd., Guangzhou 510080, China. }%
    \thanks{$^{3}$Han Ding is also with the HUST-Wuxi Research Institute, Wuxi 214174, China. }%    
\thanks{Digital Object Identifier (DOI): see top of this page.}
}

\begin{document}

\maketitle
\thispagestyle{empty}
\pagestyle{empty}

%%%%%%%%%%%%%%%%%%%%%%%%%%%%%%%%%%%%%%%%%%%%%%%%%%%%%%%%%%%%%%%%%%%%%%%%%%%%%%%%
\begin{abstract}
This paper studies tactile-only 6-DoF pose refinement and belief maintenance for grasped objects in static and short quasi-static in-hand configurations where vision is unavailable or heavily occluded. The key difficulty is tactile partial observability: whole-hand taxel contacts are sparse, intermittent, and ambiguous under limited excitation and object symmetries. We propose a physics-informed particle filter on $\mathrm{SE}(3)$ that updates pose beliefs from dense whole-hand tactile measurements. The likelihood combines active-contact signed-distance consistency, force-normal alignment, friction-cone feasibility, zero-force negative evidence, and optional feasibility guards. A sliding-window log-likelihood fuses recent tactile frames to reduce single-frame ambiguity, while a potential-field-guided proposal steers particles away from hand--object penetration. Symmetry-aware resampling preserves multiple plausible modes. Experiments on an Allegro Hand V5 with five objects show lower normalized ADD-S than tactile-only geometric, particle-filter, and learning baselines, and ablations confirm the benefits of temporal fusion, potential guidance, and mode preservation.
\end{abstract}


%%%%%%%%%%%%%%%%%%%%%%%%%%%%%%%%%%%%%%%%%%%%%%%%%%%%%%%%%%%%%%%%%%%%%%%%%%%%%%%%
% =========================
% Section I: Introduction
% =========================

\section{Introduction}
\label{sec:introduction}

\IEEEPARstart{R}{eliable}
6-DoF object pose estimation during grasping is essential for regrasping, reorientation, and contact-rich assembly. Vision pipelines become fragile when the hand occludes the object, viewpoints are limited, or RGB(-D) degrades under specularities and motion blur; consequently, strong vision-only estimators~\cite{xiangPoseCNNConvolutionalNeural2018b,wangDenseFusion6DObject2019a} may violate their sensing assumptions once an object is grasped.

Tactile sensing remains available under severe hand-object occlusion and has been widely used for contact-rich manipulation and in-hand state estimation~\cite{yousefTactileSensingDexterous2011,luoRoboticTactilePerception2017a,chalonOnlineInhandObject2013a,bauzaTac2PoseTactileObject2023a}. However, tactile-only 6-DoF pose estimation remains partially observable because active contacts are sparse, local, intermittent, and often symmetric. Existing tactile and visuo-tactile methods address related goals under different sensing assumptions and update mechanisms. Some rely on local high-resolution tactile images or first-touch contact geometry, some use sparse contact or proprioceptive cues, and visuo-tactile approaches assume additional visual/RGB-D observations. In contrast, we focus on tactile-only quasi-static pose refinement using distributed whole-hand taxel arrays with 3D contact-force measurements.

In this paper, we propose a physics-informed sliding-window particle filter for tactile-only in-hand 6-DoF pose refinement. Although it estimates a posterior on $\mathrm{SE}(3)$, the evaluated regime is low-rate refinement and belief maintenance from identity or previous-frame initialization, not high-speed global localization or dynamic tracking. The likelihood combines object--contact geometry, force--normal alignment, friction feasibility, and non-penetration evidence. A short temporal window reduces single-frame ambiguity, while potential-guided propagation steers particles away from penetration. Our contribution is the integration of these elements for distributed whole-hand taxels with measured 3D forces. The main contributions are:

\begin{itemize}
    \item A physics-informed Bayesian filtering formulation for tactile-only in-hand 6-DoF pose refinement, with a likelihood that jointly encodes geometric, normal, frictional, and non-penetration feasibility.
    \item A sliding-window particle filtering algorithm on $\mathrm{SE}(3)$ for robust belief updates under intermittent contact and noisy whole-hand tactile observations.
    \item A potential-field-guided particle propagation strategy that uses finite-difference gradients to steer particles away from implausible hand--object penetrations, improving sample efficiency and stabilizing convergence.
\end{itemize}

%%%%%%%%%%%%%%%%%%%%%%%%%%%%%%%%%%%%%%%%%%%%%%%%%%%%%%%%%%%%%%%%%%%%%%%%%%%%%%%%
%%%%%%%%%%%%%%%%%%%%%%%%%%%%%%%%%%%%%%%%%%%%%%%%%%%%%%%%%%%%%%%%%%%%%%%%%%%%%%%%
%%%%%%%%%%%%%%%%%%%%%%%%%%%%%%%%%%%%%%%%%%%%%%%%%%%%%%%%%%%%%%%%%%%%%%%%%%%%%%%%

\section{Related Work}
\label{sec:related_work}

\subsection{Tactile and visuo-tactile pose estimation}
Tactile sensing supports contact-rich manipulation and in-hand state estimation~\cite{yousefTactileSensingDexterous2011,luoRoboticTactilePerception2017a}. Early tactile-only methods used touch, proprioception, or joint torques~\cite{chalonOnlineInhandObject2013a,alvarezTactilebasedInhandObject2018a}; Bimbo et al.~\cite{bimbo2016covariance} matched tactile and local object-geometry covariances. Other methods infer pose from first touch or short local tactile observations~\cite{villalongaTactileObjectPose2021b,bauzaTac2PoseTactileObject2023a}.

Richer local optical-tactile sensing enables collision-aware estimation and detailed surface measurements~\cite{caddeoCollisionawareInhand6D2023b}. Representative high-resolution sensors, including GelSight, GelSlim, DIGIT, and TacTip, provide dense local geometry but differ from our distributed force-taxel setting~\cite{yuanGelSightHighresolutionRobot2017a,donlonGelSlimHighresolutionCompact2018a,lambetaDIGITNovelDesign2020a,ward-cherrierTacTipFamilySoft2018}. Visuo-tactile methods add visual/RGB-D cues for pose, tracking, shape, or contact inference~\cite{dikhaleVisuoTactile6DPose2022a,liViHOPEVisuotactileInhand2023,liuEnhancingGeneralizable6D2024a,leeViTaSCOPEVisuotactileImplicit2025b}, while related work uses tactile feedback for extrinsic contact reasoning~\cite{kimSimultaneousTactileEstimation2023}.

Closest to our setting are Chalon et al., Alvarez et al., and Bimbo et al.~\cite{chalonOnlineInhandObject2013a,alvarezTactilebasedInhandObject2018a,bimbo2016covariance}, but they use sparse contact/proprioception or frame-wise geometry. We instead use distributed whole-hand taxels with 3D forces and temporal Bayesian updates; unlike first-touch/local rendering or visuo-tactile methods, our setting is tactile-only whole-hand pose refinement.

\subsection{Contact-aware Bayesian estimation and filtering}
Contact-rich localization is also posed as probabilistic inference under geometric/physical constraints. Particle and manifold particle filters support touch/contact localization~\cite{petrovskayaGlobalLocalizationObjects2011a,kovalPoseEstimationContact2013a,kovalManifoldParticleFilter2017}. Related tools include Monte Carlo localization, smoothing, pose uncertainty, invariant/contact-aided filtering~\cite{dellaertMonteCarloLocalization1999,kaessISAM2IncrementalSmoothing2012,barfootAssociatingUncertaintyThreedimensional2014,barrauInvariantExtendedKalman2017,barrauInvariantKalmanFiltering2018,hartleyContactaidedInvariantExtended2020}, and EKF-style in-hand/contact estimators~\cite{pfanneEKFbasedInhandObject2017a,siposSimultaneousContactLocation2022a,vandermerweSimultaneousExtrinsicContact2026}.

Unlike contact-localization particle filters~\cite{petrovskayaGlobalLocalizationObjects2011a,kovalPoseEstimationContact2013a,kovalManifoldParticleFilter2017}, we refine a grasped-object pose from whole-hand taxels and measured 3D forces using contact SDF, force--normal, friction, zero-force, and auxiliary feasibility evidence over a short window. Unlike invariant/contact-aided filters~\cite{barrauInvariantExtendedKalman2017,barrauInvariantKalmanFiltering2018,hartleyContactaidedInvariantExtended2020}, our state is object rather than robot-body pose; unlike EKF variants~\cite{pfanneEKFbasedInhandObject2017a,siposSimultaneousContactLocation2022a,vandermerweSimultaneousExtrinsicContact2026}, particles retain symmetry-induced modes on $\mathrm{SE}(3)$. The novelty is this integrated formulation rather than an isolated residual.

%%%%%%%%%%%%%%%%%%%%%%%%%%%%%%%%%%%%%%%%%%%%%%%%%%%%%%%%%%%%%%%%%%%%%%%%%%%%%%%%
%%%%%%%%%%%%%%%%%%%%%%%%%%%%%%%%%%%%%%%%%%%%%%%%%%%%%%%%%%%%%%%%%%%%%%%%%%%%%%%%
%%%%%%%%%%%%%%%%%%%%%%%%%%%%%%%%%%%%%%%%%%%%%%%%%%%%%%%%%%%%%%%%%%%%%%%%%%%%%%%%

\section{Method}
\label{sec:method}

The method contains three core components: a physics-informed tactile likelihood, sliding-window particle-filter updates on $\mathrm{SE}(3)$, and potential-field-guided particle propagation. Contact-required, front-side, quasi-static equilibrium, symmetry-preserving resampling, adaptive proposal scaling, and relocalization are auxiliary robustness mechanisms rather than standalone contributions.

\subsection{Problem Setup}
We consider a multi-finger dexterous hand equipped with dense tactile taxels. At each time step $t$, the hand provides a set of taxel positions and 3D contact forces expressed in the world frame $\mathcal{F}_W$. Let $\mathcal{F}_O$ denote the object frame, and let ${}^W T_O\in \mathrm{SE}(3)$ denote the object pose in the world frame as
\begin{equation}
{}^W T_O \triangleq
\begin{bmatrix}
\mathbf{R} & \mathbf{t}\\
\mathbf{0}^\top & 1
\end{bmatrix},
\qquad
\mathbf{R}\in \mathrm{SO}(3),\ \mathbf{t}\in\mathbb{R}^3,
\label{eq:wto}
\end{equation}
where the rotation $\mathbf{R}$ is parameterized by a unit quaternion $\mathbf{q}\in\mathbb{H}$. A point $\mathbf{p}^W$ expressed in the world frame transforms into the object frame as
\begin{equation}
\mathbf{p}^O = \mathbf{R}^\top(\mathbf{p}^W-\mathbf{t}).
\label{eq:pw_to_po}
\end{equation}

At time $t$, the tactile observation is written as
\begin{equation}
\label{eq:tactile_observation}
\mathcal{Z}_t \triangleq
\Big\{\big(\mathbf{p}_k^W, \mathbf{f}_k^W, \mathbf{z}_k^W\big)\Big\}_{k=1}^{K},
\end{equation}
where $\mathbf{p}_k^W$ is the center of the $k$-th taxel in $\mathcal{F}_W$, $\mathbf{f}_k^W$ is the measured 3D force after sign correction so that it is consistent with outward object normals, and $\mathbf{z}_k^W$ is the outward front direction of the taxel, which is later used in a half-space feasibility constraint. We partition taxels into active-contact and zero-force sets using thresholds $\tau_c$ and $\tau_0$ as
\begin{equation}
\label{eq:active_zero_sets}
\mathcal{C}_t \triangleq \{k: \|\mathbf{f}_k^W\|>\tau_c\},
\qquad
\mathcal{Z}_t^{0} \triangleq \{k: \|\mathbf{f}_k^W\|\le \tau_0\}.
\end{equation}

Given an object mesh $\mathcal{M}$, we assume access to two geometric operators defined in the object frame. The first is the signed distance field (SDF) $d:\mathbb{R}^3\rightarrow\mathbb{R}$, which is positive outside the object and negative inside. The second is a nearest-point operator returning the closest surface point and the corresponding outward unit normal as
\begin{equation}
\label{eq:mesh_nn}
(\mathbf{q}^O(\mathbf{p}^O),\,\mathbf{n}^O(\mathbf{p}^O)) = \mathrm{NN}_{\mathcal{M}}(\mathbf{p}^O).
\end{equation}

For notational convenience, we also use
\begin{equation}
\label{eq:world_normal}
\mathbf{n}^W(\mathbf{p}^O) \triangleq \mathbf{R}\,\mathbf{n}^O(\mathbf{p}^O).
\end{equation}
This is the corresponding normal expressed in the world frame.

\subsection{Sliding-Window Contact Particle Filter}
\label{subsec:pf}

%%%%%%%%%%%%%%%%%%%%%%%%%%%%%%%%%%%%%%%%%
\begin{table*}[!t]
\centering
\caption{Key full-model parameters. Superscript i/f denotes initial/final values.}
\label{tab:method_params}
\scriptsize
\setlength{\tabcolsep}{2.6pt}
\renewcommand{\arraystretch}{0.86}
\begin{tabular*}{\textwidth}{@{\extracolsep{\fill}}llllll@{}}
\toprule
Param. & Value & Param. & Value & Param. & Value \\
\midrule
$\tau_c$ & $0.5$ &
$F_{\rm ref},\omega_{\max}$ & $1.0$, $14.0$ &
$\sigma_{\rm sd},\lambda_{\rm sd}$ & $0.002$, $6.0$ \\

$c_d,\delta_{\min},\delta_{\max}$ & $0.0018$, $0$, $0.007$ &
$\sigma_{\rm ang},\lambda_{\rm ang}$ & $25^\circ$, $1.0$ &
$\mu,\sigma_{\rm fric},\lambda_{\rm fric}$ & $0.6$, $1.0$, $1.0$ \\

$m_0,\sigma_0,\lambda_0$ & $0.001$, $0.002$, $1.0$ &
$m_f,\sigma_{\rm front},\lambda_{\rm front}$ & $0.001$, $0.002$, $10.0$ &
$\epsilon_{\rm out},\lambda_{\rm out}$ & $0.006$, $1.0$ \\

$\sigma_g,\lambda_F,\lambda_\tau$ & $0.002$, $0.5$, $0.5$ &
$N$ & $500$ &
$\sigma_{\parallel}^{\mathrm{i/f}}$ & $0.03/0.002\,\mathrm{m}$ \\

$\sigma_{\perp}^{\mathrm{i/f}}$ & $0.01/0.001\,\mathrm{m}$ &
$\sigma_{\rm yaw}^{\mathrm{i/f}}$ & $10.0/0.8^\circ$ &
$\sigma_{\rm tilt}^{\mathrm{i/f}}$ & $3.0/0.04^\circ$ \\

$N_a$ & $200$ &
$\phi(\cdot)$ & piecewise, $k=6.0$, hold $0.2$ &
$\gamma_{\rm explore}$ & $0.10$ \\

$\kappa_{\rm explore}$ & $2.0$ &
$s_{\min},s_{\max}$ & $0.25$, $2.0$ &
$\tau_{\rm ess}$ & $0.05$ \\

$\beta$ & $0.4$ &
$W,\alpha$ & $5$, $0.7$ &
$m_{\rm pot},N_{\rm pot}$ & $0.0005\,\mathrm{m}$, $256$ \\

$\rho_{\rm guide},I_{\rm guide}$ & $0.30$, $2$ &
$\eta_t,\eta_r$ & $0.05$, $0.06$ &
$\epsilon_t,\epsilon_r$ & $10^{-3}\,\mathrm{m}$, $10^{-3}\,\mathrm{rad}$ \\

$\Delta t_{\max},\Delta\theta_{\max},h_{\rm slab}$ &
$0.005\,\mathrm{m}$, $0.05\,\mathrm{rad}$, $0.03\,\mathrm{m}$ &
\multicolumn{4}{c}{} \\
\bottomrule
\end{tabular*}
\vspace{-10pt}
\end{table*}
%%%%%%%%%%%%%%%%%%%%%%%%%%%%%%%%%%%%%%%%%

We estimate the posterior $p(\mathbf{x}_t\mid \mathcal{Z}_{1:t})$ over the object pose as
\begin{equation}
\mathbf{x}_t \triangleq (\mathbf{t}_t,\mathbf{q}_t).
\label{eq:pose_state}
\end{equation}
using a particle filter with $N$ weighted particles as
\begin{equation}
\{(\mathbf{x}_t^{(i)}, w_t^{(i)})\}_{i=1}^N.
\label{eq:particle_set}
\end{equation}

We monitor the effective sample size (ESS) of the normalized particle weights as
\begin{equation}
\mathrm{ESS}_t =
\frac{1}{\sum_{i=1}^{N}(w_t^{(i)})^2},
\qquad
r_{\mathrm{ESS},t}=\frac{\mathrm{ESS}_t}{N}.
\label{eq:rw_proposal}
\end{equation}
Here $r_{\mathrm{ESS},t}$ denotes the ESS ratio.

The filter combines stochastic propagation on $\mathrm{SE}(3)$, short-horizon temporal fusion, and a physics-informed tactile likelihood. At each iteration, particle propagation starts from a random-walk proposal on $\mathrm{SE}(3)$ as
\begin{equation}
\mathbf{x}_t^{(i)} \sim p(\mathbf{x}_t\mid \mathbf{x}_{t-1}^{(i)}),
\qquad
\mathbf{x}=(\mathbf{t},\mathbf{q}).
\end{equation}

The proposal variance is scheduled by an annealing factor so that exploration is stronger in the early stage of refinement and gradually decreases as the posterior concentrates. Let $n$ denote the current iteration index and let $N_a$ denote the annealing horizon. We define
\begin{equation}
\tilde{t}=\mathrm{clip}\!\left(\frac{n}{N_a},\,0,\,1\right),
\qquad
g = \phi(\tilde{t})\in[0,1],
\end{equation}
where $\phi(\cdot)$ is a configurable schedule, such as a linear, exponential, or piecewise decay. For any noise scale $\sigma$, we interpolate between an initial and a final value according to
\begin{equation}
\sigma(g)= (1-g)\,\sigma_{\mathrm{init}} + g\,\sigma_{\mathrm{final}}.
\label{eq:anneal_interp}
\end{equation}

To better match the local geometry of in-hand refinement, we optionally use anisotropic perturbations aligned with a selected object axis. Let $\mathbf{a}^W$ be a chosen object axis expressed in the world frame and computed from the current particle orientation, and let $(\mathbf{a}^W,\mathbf{b}^W,\mathbf{c}^W)$ form an orthonormal basis. When anisotropic translation is enabled, we sample
\begin{equation}
\begin{aligned}
\Delta \mathbf{t}
&= \epsilon_{\parallel}\mathbf{a}^W
+ \epsilon_b\mathbf{b}^W
+ \epsilon_c\mathbf{c}^W,\\
\epsilon_{\parallel}&\sim\mathcal{N}(0,\sigma_{\parallel}^2),\qquad
\epsilon_b,\epsilon_c\sim\mathcal{N}(0,\sigma_{\perp}^2),
\end{aligned}
\label{eq:trans_aniso}
\end{equation}
where $(\sigma_{\parallel},\sigma_{\perp})$ are obtained from Eq.~\eqref{eq:anneal_interp}. Rotation is perturbed in the same basis by sampling a small increment as
\begin{equation}
\begin{aligned}
\boldsymbol{\omega}
&= \delta\psi\,\mathbf{a}^W + \delta\theta_b\,\mathbf{b}^W + \delta\theta_c\,\mathbf{c}^W,\\
\delta\psi&\sim\mathcal{N}(0,\sigma_{\mathrm{yaw}}^2),\qquad
\delta\theta_b,\delta\theta_c\sim\mathcal{N}(0,\sigma_{\mathrm{tilt}}^2),
\end{aligned}
\label{eq:rot_aniso}
\end{equation}
where $(\sigma_{\mathrm{yaw}},\sigma_{\mathrm{tilt}})$ are scheduled in the same way. The orientation is then updated on $\mathrm{SO}(3)$ by left-multiplication as
\begin{equation}
\mathbf{R} \leftarrow \exp([\boldsymbol{\omega}]_{\times})\,\mathbf{R},
\qquad
\mathbf{q}\leftarrow \mathrm{Exp}(\boldsymbol{\omega})\otimes\mathbf{q}.
\label{eq:orientation_update}
\end{equation}

\textit{Proposal parameterization.}
The proposal in Eqs.~\eqref{eq:trans_aniso}--\eqref{eq:rot_aniso} is used only for propagation. The dominant mesh principal/box axis $\mathbf{a}^O$ is transformed as $\mathbf{a}^W=\mathbf{R}\mathbf{a}^O$; elongated objects use larger perturbations along/about it, whereas compact objects use isotropic scales. Initial scales reflect pose-prior uncertainty, final scales the desired refinement resolution, and $N_a$ the available update budget; the shared values are in Table~\ref{tab:method_params}.

The adaptive multiplier is deterministic. Let $\ell_{\rm best,j}=\max_i\ell_{\rm SW}^{(i)}$, $c_j=\lambda_{\max}(A_j)$ from Eq.~\eqref{eq:quaternion_mean_matrix}, $\Delta\ell_t=\ell_{\rm best,t-1}-\ell_{\rm best,t-2}$, and $\Delta c_t=c_{t-1}-c_{t-2}$. Set $s_t=1$ for the first two iterations. Thereafter, $s_t=s_{\max}$ if $r_{\rm ESS,t-1}<\tau_{\rm ess}$ and either increment is nonpositive; $s_t=s_{\min}$ if $r_{\rm ESS,t-1}\ge\tau_{\rm ess}$ and both are positive; otherwise $s_t=1$. A random fraction $\gamma_{\rm explore}$ uses $\kappa_{\rm explore}s_t$ to retain exploration. This controls the next proposal, separately from current-step ESS resampling.

\subsection{Potential-Guided Proposal and Sliding-Window Update}
\label{subsec:multimodal}

Before random diffusion, we apply a proposal-level hand--object potential drift to the top-weight particles. For a particle pose $X=(\mathbf{R},\mathbf{t})$, object surface samples $\mathbf{p}_j^O$ are transformed as $\mathbf{R}\mathbf{p}_j^O+\mathbf{t}$. Given the hand signed distance $d_H(\cdot)$, with negative values indicating penetration, we define
\begin{equation}
U(X)=\frac{1}{|\mathcal{S}|}
\sum_{\mathbf{p}_j^O\in\mathcal{S}}
\left[
\max\big(0,m_{\mathrm{pot}}-d_H(\mathbf{R}\mathbf{p}_j^O+\mathbf{t})\big)
\right]^2 ,
\label{eq:hand_object_potential}
\end{equation}
where $\mathcal{S}$ is a set of sampled object surface points, optionally restricted to the current contact-region slab. We estimate $\nabla_{\mathbf{t}}U$ and $\nabla_{\boldsymbol{\omega}}U$ by central finite differences and apply
\begin{equation}
\mathbf{t}_i \leftarrow \mathbf{t}_i-\eta_t\,\mathrm{clip}_t(\nabla_{\mathbf{t}}U_i),~
\mathbf{R}_i \leftarrow
\exp\!\left(-\eta_r[\mathrm{clip}_r(\nabla_{\boldsymbol{\omega}}U_i)]_{\times}\right)\mathbf{R}_i .
\label{eq:potential_drift_update}
\end{equation}

This drift is applied every two particle-filter iterations to the highest-weight 30\% particles using 256 surface samples; all proposal and drift parameters are reported in Table~\ref{tab:method_params}.

Instead of evaluating a pose hypothesis using only the current tactile frame, we maintain a FIFO buffer of the most recent $W$ observations and evaluate each particle over a short sliding window. Let $B_t=\min(W,t)$ and let $\alpha\in(0,1]$ be an exponential forgetting factor. For a pose hypothesis $\mathbf{x}$, we define the sliding-window score as
\begin{equation}
\ell_{\mathrm{SW}}(\mathbf{x})
\triangleq
\sum_{k=0}^{B_t-1}
\alpha^k
\ell(\mathbf{x};\mathcal{Z}_{t-k}),~
\log p(\mathcal{Z}_{t-B_t+1:t}|\mathbf{x})
\propto
\ell_{\mathrm{SW}}(\mathbf{x}).
\label{eq:sliding_window_score}
\end{equation}

For particle $i$, we write
$\ell_{\mathrm{SW}}^{(i)}
=
\ell_{\mathrm{SW}}(\mathbf{x}_t^{(i)})$.
Here, $\ell(\mathbf{x};\mathcal{Z}_{t-k})$ denotes the per-frame log-likelihood introduced in the next subsection. This short-horizon temporal fusion improves robustness to intermittent contact changes and brief sensing dropouts while avoiding excessive accumulation of stale evidence.

Given the sliding-window score $\ell_{\mathrm{SW}}^{(i)}$ for particle $i$, we update the particle weights using a tempered likelihood as
\begin{equation}
\tilde{w}_t^{(i)} = \exp\!\big(\beta\,\ell_{\mathrm{SW}}^{(i)}\big),
\qquad
w_t^{(i)} = \frac{\tilde{w}_t^{(i)}}{\sum_{j=1}^{N}\tilde{w}_t^{(j)}},
\label{eq:weight_update}
\end{equation}
where $\beta\in(0,1]$ controls the sharpness of the weight distribution. After normalization, the ESS ratio defined above is used to detect weight degeneracy and trigger resampling.

When $r_{\mathrm{ESS},t}<\tau_{\mathrm{ess}}$, low-variance resampling is performed to avoid weight degeneracy. Algorithm~\ref{alg:swcpf} summarizes the propagation, windowed scoring, ESS-based resampling, optional mode preservation, and relocalization checks.

%%%%%%%%%%%%%%%%%%%%%%%%%%%%%%%%%%%%%%%%%%%%%%%%%%%%%%%%%%%%%%%%%%%%%%%%%%%%%%%%%%%
\begin{algorithm}[!t]
\caption{Sliding-Window Contact Particle Filter}
\label{alg:swcpf}
\footnotesize
\begin{algorithmic}[1]
\REQUIRE $\mathcal{Z}_t$, geometry $(\mathcal{M},d,\mathrm{NN}_{\mathcal{M}})$, particles $\{(\mathbf{x}_{t-1}^{(i)},w_{t-1}^{(i)})\}_{i=1}^N$
\ENSURE Estimated pose $\hat{\mathbf{x}}_t$
\STATE Partition taxels by Eq.~\eqref{eq:active_zero_sets}; push $\mathcal{Z}_t$ into the FIFO window and set $B_t=\min(W,t)$.
\STATE Set proposal scale $s_t$ from previous diagnostics; optionally apply Eq.~\eqref{eq:potential_drift_update} to top-weight particles.
\FOR{$i=1,\ldots,N$}
    \STATE Sample $\mathbf{x}_t^{(i)}$ using Eqs.~\eqref{eq:trans_aniso}--\eqref{eq:rot_aniso} with $s_t$; use $\kappa_{\rm explore}s_t$ for a random $\gamma_{\rm explore}$ fraction.
    \STATE Compute $\ell_{\rm SW}^{(i)}$ by Eq.~\eqref{eq:sliding_window_score} and $\tilde{w}_t^{(i)}$ by Eq.~\eqref{eq:weight_update}.
\ENDFOR
\STATE Normalize weights; compute $r_{\rm ESS,t}$, $\ell_{\rm best}$, and $\bar d_{\rm pen}$.
\IF{the relocalization criterion is satisfied}
    \STATE Reinitialize by Eqs.~\eqref{eq:coarse_contact_pose}--\eqref{eq:coarse_translation} and reset $w_t^{(i)}=1/N$.
\ELSIF{$r_{\rm ESS,t}<\tau_{\rm ess}$}
    \STATE Resample, optionally grouped by the mode label in Eq.~\eqref{eq:symmetry_mode_label}.
\ENDIF
\STATE Return the dominant-mode weighted estimate $\hat{\mathbf{x}}_t$.
\end{algorithmic}
\end{algorithm}
%%%%%%%%%%%%%%%%%%%%%%%%%%%%%%%%%%%%%%%%%%%%%%%%%%%%%%%%%%%%%%%%%%%%%%%%%%%%%%%%%%%

\subsection{Likelihood Factors as Robust Soft Constraints}
\label{subsec:likelihood}

The observation model is a sum of robust soft-constraint penalties in log space,
\begin{equation}
\ell(\mathbf{x};\mathcal{Z}_t)=\sum_m \ell_m(\mathbf{x};\mathcal{Z}_t),
\quad \ell_m(\cdot)\le 0 ,
\label{eq:generic_likelihood}
\end{equation}
where larger values indicate smaller inconsistency. Eqs.~\eqref{eq:normalized_residual}--\eqref{eq:generic_factor} define a residual-agnostic Huber penalty template; the following paragraphs instantiate it for contact SDF, force-normal, friction-cone, zero-force, front-side, and quasi-static-equilibrium residuals. The expanded likelihood in Eq.~\eqref{eq:full_likelihood} is the concrete implementation of this template.

Each factor is defined from either a scalar residual $r$ or a vector residual $\mathbf{r}$, together with a scale $\sigma>0$ and a weight $\lambda>0$. We use the normalized magnitude as
\begin{equation}
\bar{r} \triangleq \frac{|r|}{\sigma}
\qquad\text{or}\qquad
\bar{r} \triangleq \frac{\|\mathbf{r}\|_2}{\sigma},
\label{eq:normalized_residual}
\end{equation}
together with the Huber loss
\begin{equation}
\rho_{\delta}(\bar{r}) =
\begin{cases}
\frac{1}{2}\bar{r}^2, & \bar{r}\le \delta,\\
\delta\bigl(\bar{r}-\frac{1}{2}\delta\bigr), & \bar{r}>\delta.
\end{cases}
\label{eq:huber_loss}
\end{equation}

For active-contact taxels, where the index $k$ denotes an individual contact taxel/contact sample with $k\in\mathcal{C}_t$, we define bounded per-contact weights based on force magnitude as
\begin{equation}
\omega_k = \mathrm{clip}\!\left(\frac{\|\mathbf{f}_k^W\|}{F_{\mathrm{ref}}},\,0,\,\omega_{\max}\right).
\label{eq:contact_weight}
\end{equation}

Each factor is then written in the generic form as
\begin{equation}
\ell_m = -\lambda \sum_{k} \omega_k\,\rho_{\delta}(\bar{r}_k).
\label{eq:generic_factor}
\end{equation}
Here $k$ refers to the taxel/sample associated with each residual: $k\in\mathcal{C}_t$ for active-contact factors and $k\in\mathcal{Z}_t^0$ for the zero-force term.

This common formulation allows multiple geometric and physical consistency terms to be combined into a single robust tactile likelihood.

\paragraph{Signed-distance residual}
For each active-contact taxel $k\in\mathcal{C}_t$, we first enforce signed-distance consistency between the measured taxel position and the object surface. After transforming $\mathbf{p}_k^W$ into $\mathbf{p}_k^O$ using Eq.~\eqref{eq:pw_to_po}, we evaluate the signed distance $d_k=d(\mathbf{p}_k^O)$. To account for small compliant indentation, we model a force-dependent offset as
\begin{equation}
\delta_k = \mathrm{clip}\!\big(c_d\,\|\mathbf{f}_k^W\|,\,\delta_{\min},\,\delta_{\max}\big),
\label{eq:indentation_offset}
\end{equation}
where $c_d$ is a depth-per-Newton coefficient and $(\delta_{\min},\delta_{\max})$ bound the expected indentation range. The resulting residual is
\begin{equation}
r^{\mathrm{sd}}_k = d_k + \delta_k \approx 0
\qquad
(\text{target: } d_k \approx -\delta_k),
\label{eq:contact_sd_res}
\end{equation}
and the corresponding factor $\ell_{\mathrm{sd}}$ follows Eq.~\eqref{eq:generic_factor}. Optionally, we add a contact-required penalty when a forceful taxel lies far outside the object:
\begin{equation}
\ell_{\mathrm{req}} = -\lambda_{\mathrm{out}} \sum_{k\in\mathcal{C}_t} \mathbf{1}[d_k>\epsilon_{\mathrm{out}}].
\label{eq:contact_required}
\end{equation}

The optional guard $\ell_{\mathrm{req}}$ is enabled only for forceful active taxels; $\epsilon_{\mathrm{out}}$ is set above the expected tactile localization and mesh-query errors so that it penalizes only grossly inconsistent outside-surface contacts.

\paragraph{Force-normal residual}
For each active taxel, we use the nearest surface normal to measure
force-normal agreement,
\begin{equation}
(\mathbf{q}_k^O,\mathbf{n}_k^O)=\mathrm{NN}_{\mathcal{M}}(\mathbf{p}_k^O),
\qquad
\mathbf{n}_k^W=\mathbf{R}\mathbf{n}_k^O,
\label{eq:active_surface_normal}
\end{equation}
and let $\hat{\mathbf{f}}_k^W=\mathbf{f}_k^W/\|\mathbf{f}_k^W\|$. We define the angular residual as
\begin{equation}
r^{\mathrm{ang}}_k =
\arccos\!\Big( \mathrm{clip}\big((\hat{\mathbf{f}}_k^W)^\top \mathbf{n}_k^W,\,-1,\,1\big)\Big),
\label{eq:angle_res}
\end{equation}
and compute the factor $\ell_{\mathrm{ang}}$ using the same robust form.

\paragraph{Friction-cone residual}
Friction feasibility is imposed by decomposing the force into normal and tangential components,
\begin{equation}
\begin{aligned}
F_{n,k} &= \max\big(0, (\mathbf{f}_k^W)^\top \mathbf{n}_k^W\big),\\
\mathbf{f}_{t,k} &= \mathbf{f}_k^W - \big((\mathbf{f}_k^W)^\top \mathbf{n}_k^W\big)\mathbf{n}_k^W,\\
F_{t,k} &= \|\mathbf{f}_{t,k}\|_2.
\end{aligned}
\label{eq:force_decomposition}
\end{equation}

The corresponding friction-cone violation is
\begin{equation}
r^{\mathrm{fric}}_k = \max\big(0,\,F_{t,k}-\mu F_{n,k}\big).
\label{eq:fric_res}
\end{equation}
It yields $\ell_{\mathrm{fric}}$ through Eq.~\eqref{eq:generic_factor}. The coefficient $\mu$ is a conservative nominal value for soft tactile contact rather than an online estimate.

\paragraph{Front-side feasibility residual}

The taxel front direction provides an auxiliary half-space guard against backside contacts. Let $\mathbf{q}_k^W = \mathbf{t}+\mathbf{R}\mathbf{q}_k^O$ denote the closest surface point expressed in the world frame. We define
\begin{equation}
s_k = (\mathbf{q}_k^W-\mathbf{p}_k^W)^\top \mathbf{z}_k^W,  r^{\mathrm{front}}_k = \max(0,\,m_f - s_k),
\label{eq:front_res}
\end{equation}
where $m_f \ge 0$ is a margin, which yields the factor $\ell_{\mathrm{front}}$. The margin $m_f$ is selected according to the taxel surface thickness and localization uncertainty; it is used only to reject backside contacts that are incompatible with the known taxel front direction.

\paragraph{Zero-force residual}
Besides positive evidence from active contacts, we also use negative evidence from zero-force taxels. For each $k\in\mathcal{Z}_t^{0}$, we penalize pose hypotheses for which the taxel center lies inside or too close to the object as
\begin{equation}
r^{0}_k = \max\big(0,\,m_0 - d(\mathbf{p}_k^O)\big),
\label{eq:zero_sdf_res}
\end{equation}
where $m_0>0$ is a safety margin. This produces the factor $\ell_{0}$ and helps suppress implausible hypotheses that would predict contact where no contact is observed.

\paragraph{Static-equilibrium residual}
As an auxiliary weak check under quasi-static interaction, we gate each
measured force by its contact-SDF consistency before evaluating global
force/torque balance. For $k\in\mathcal{C}_t$,
\begin{equation}
e_k \triangleq d(\mathbf{p}_k^O)+\delta_k,~
g_k \triangleq \exp\!\Big(-\tfrac{1}{2}(e_k/\sigma_g)^2\Big)\in(0,1],
\label{eq:equilibrium_gate}
\end{equation}
where $\sigma_g$ is a gating scale, which controls how quickly off-surface forceful taxels are down-weighted. The gated net force and torque about the world origin are then
\begin{equation}
\mathbf{F} = \sum_{k\in\mathcal{C}_t} g_k\,\mathbf{f}_k^W,
\qquad
\boldsymbol{\tau}_0 = \sum_{k\in\mathcal{C}_t} g_k\,(\mathbf{p}_k^W \times \mathbf{f}_k^W).
\label{eq:gated_wrench}
\end{equation}

Let the object mass be $m$, let gravity be $\mathbf{g}^W$, and let the object center of mass in the world frame be
\begin{equation}
\mathbf{c}^W=\mathbf{t}+\mathbf{R}\mathbf{c}^O,
\label{eq:world_com}
\end{equation}
where $\mathbf{c}^O$ is known from CAD and defaults to $\mathbf{0}$. We then define the residuals
\begin{equation}
\mathbf{r}^{F} = \mathbf{F} + m\mathbf{g}^W,
\qquad
\mathbf{r}^{\tau_g} = \boldsymbol{\tau}_0 + \mathbf{c}^W \times (m\mathbf{g}^W),
\label{eq:equilibrium_residual}
\end{equation}
and convert them into log-likelihood terms using \eqref{eq:generic_factor}, yielding
\begin{equation}
\ell_{\mathrm{eq}}=\ell_{F}+\ell_{\tau_g}.
\label{eq:equilibrium_likelihood}
\end{equation}

Combining the instantiated factors above, Eq.~\eqref{eq:generic_likelihood} becomes the following concrete per-frame log-likelihood used in our implementation
\begin{equation}
\ell(\mathbf{x};\mathcal{Z}_t) =
\ell_{\mathrm{sd}} + \ell_{\mathrm{ang}} + \ell_{\mathrm{fric}}
+ \ell_{0} + \ell_{\mathrm{front}} + \ell_{\mathrm{eq}} + \ell_{\mathrm{req}}.
\label{eq:full_likelihood}
\end{equation}

In Eq.~\eqref{eq:full_likelihood}, $\ell_{\mathrm{sd}}$, $\ell_{\mathrm{ang}}$, $\ell_{\mathrm{fric}}$, and $\ell_0$ are the main physics-informed terms; the others are auxiliary guards for gross outside-surface contacts, backside contacts, and weak quasi-static equilibrium. They are enabled only when the required sensing quantity or assumption is available. For transfer, contact thresholds are set above no-contact bias/noise; distance scales and margins follow taxel localization uncertainty, SDF resolution, and compliance; angular scales follow calibrated force-direction dispersion; and physical constants use conservative nominal values. Factor weights are chosen once on held-out validation sequences so no normalized factor dominates, then fixed across all objects and tests.

\subsection{Multi-modality Preservation for Symmetric Objects}
Symmetric objects naturally induce multi-modal posteriors because multiple object poses may explain the same local tactile pattern equally well. To reduce premature collapse to an arbitrary symmetric equivalent, we cluster particles using a symmetry-aware mode identifier built from two components: a principal-axis flip label and a discretized yaw bin around that axis. Let $\mathbf{a}_i^W$ denote the principal axis of particle $i$ in world coordinates. Given a reference axis $\mathbf{a}_{\mathrm{ref}}^W$, we define
\begin{equation}
\begin{aligned}
\mathrm{flip}_i &= \mathbf{1}\big[(\mathbf{a}_i^W)^\top \mathbf{a}_{\mathrm{ref}}^W < 0\big],\\
\mathrm{yaw}_i &= \left\lfloor \frac{\psi_i}{2\pi} B_{\psi} \right\rfloor \in \{0,\dots,B_{\psi}-1\},
\end{aligned}
\label{eq:symmetry_components}
\end{equation}
where $\psi_i$ is the yaw angle of particle $i$ about its principal axis and $B_{\psi}$ is the number of yaw bins. The resulting mode label is
\begin{equation}
\mathrm{mode}_i = \mathrm{flip}_i B_{\psi} + \mathrm{yaw}_i.
\label{eq:symmetry_mode_label}
\end{equation}

The mode label in Eq.~\eqref{eq:symmetry_mode_label} is intended for objects with one dominant rotational symmetry axis or approximate principal axis. For multiple independent symmetry axes, it could be generalized to a discrete label over the corresponding finite symmetry group, but such objects are not evaluated here; our claim is therefore restricted to single- or dominant-axis symmetry. The yaw-bin number $B_\psi$ is chosen coarse enough to avoid noise-induced mode fragmentation while still separating symmetry-related hypotheses, and is fixed across experiments.

During resampling, particles are grouped by mode so that each plausible symmetry-related hypothesis retains a minimum particle budget. Let
\begin{equation}
\mathcal{I}_g=\{i:\mathrm{mode}_i=g\},
\qquad
W_g=\sum_{i\in\mathcal{I}_g} w_i
\label{eq:mode_mass}
\end{equation}
denote the particle index set and total weight mass of mode $g$, respectively. We allocate a per-mode resample count $N_g$ proportional to $W_g$ with a small floor, and then perform low-variance resampling independently within each mode. This mode-preserving strategy maintains multiple plausible explanations rather than collapsing the entire particle set to a single symmetric equivalent.

The final pose estimate is taken from the dominant mode, namely the mode with the largest mass $W_g$. Within that mode, the translation is estimated by the weighted mean as
\begin{equation}
\hat{\mathbf{t}}=\sum w_i\mathbf{t}_i \Big/ \sum w_i,
\label{eq:translation_mean}
\end{equation}
and the quaternion mean is computed as the principal eigenvector of
\begin{equation}
A=\sum \bigl(w_i/\sum w_i\bigr)\mathbf{q}_i\mathbf{q}_i^\top,
\label{eq:quaternion_mean_matrix}
\end{equation}
after sign-consistency correction.

\subsection{Degeneracy Detection and Relocalization}

The relocalization mechanism is used as an auxiliary recovery module rather than as a regular likelihood term or a separate claimed contribution. To improve robustness against particle collapse and severe initialization error, we monitor the log weight normalizer, the ESS ratio, the best-particle score \(\ell_{\mathrm{best},t}=\max_i\ell_{\mathrm{SW}}(\mathbf{x}_t^{(i)})\), and the best-particle mean hand--object penetration depth \(\bar d_{\mathrm{pen}}\), computed over the object surface samples used for penetration checking. When severe degeneracy is detected, the filter resets the particle set around a contact-driven coarse estimate. If enabled, a few damped Gauss--Newton steps refine only this initial seed using the active-contact SDF and zero-force residuals in Sec.~\ref{subsec:likelihood}; they neither add a likelihood term nor change Eq.~\eqref{eq:full_likelihood}.

\textit{Triggering criterion.}
Ordinary resampling and relocalization use different thresholds. Resampling is triggered by \(r_{\mathrm{ESS},t}<\tau_{\mathrm{ess}}=0.05\) and does not imply relocalization. Relocalization is considered only when at least \(K_{\min}=6\) active taxels are available and either (i) the best particle has \(\bar d_{\mathrm{pen}}>0.03\,\mathrm{m}\), or (ii) \(r_{\mathrm{ESS},t}<0.005\) and \(\ell_{\mathrm{best},t}<-10^6\) persist for \(P_{\mathrm{reloc}}=4\) particle-filter iterations. This trigger starts recovery and is not an evaluation failure label by itself.

Once the trigger condition is satisfied, particles are reinitialized using the contact-driven coarse estimate in Eqs.~\eqref{eq:coarse_contact_pose}--\eqref{eq:coarse_translation}. The vector $\mathbf{u}$ in Eq.~\eqref{eq:coarse_contact_pose} denotes the normalized average contact-force direction computed from the active taxels. If enabled, the optional local SDF refinement reuses the active-contact SDF residual in Eq.~\eqref{eq:contact_sd_res} and the zero-force residual in Eq.~\eqref{eq:zero_sdf_res}. This recovery step does not modify the measurement likelihood; it only resets or locally adjusts the particle set when the current belief becomes severely inconsistent with the tactile observations.

The coarse initialization is constructed from the current active contacts. Specifically, we compute a weighted contact centroid and an average force direction as
\begin{equation}
\mathbf{c} = \frac{\sum_{k\in\mathcal{C}_t} {\omega}_k \mathbf{p}_k^W}{\sum_{k\in\mathcal{C}_t}{\omega}_k},
\qquad
\mathbf{u} = \frac{\sum_{k\in\mathcal{C}_t}\mathbf{f}_k^W}{\left\|\sum_{k\in\mathcal{C}_t}\mathbf{f}_k^W\right\|}.
\label{eq:coarse_contact_pose}
\end{equation}

To reduce the risk of initializing the object inside the hand, the translation is offset along $-\mathbf{u}$ as
\begin{equation}
\mathbf{t}_0 = \mathbf{c} - \gamma_{\rm init}\,\mathbf{u},
\label{eq:coarse_translation}
\end{equation}
where $\gamma_{\rm init}>0$ is an offset. The initial orientation is identity for an uninformative prior, or estimated from coarse contact geometry when available.

\begin{table*}[!t]
\centering
\caption{Normalized ADD-S (median/IQR) and runtime. Contact-SDF PF and Single-frame Physics PF are internal ablations; the other comparators are external baselines.}
\label{tab:baseline_expanded}
\scriptsize
\setlength{\tabcolsep}{2.8pt}
\begin{tabular}{lccccccc}
\toprule
Method & Screwdriver & Connector & Cuboid & Flat bottle & Banana & Average & Time / update (s) \\
\midrule
ICP
& 1.103/0.235 & 0.517/0.171 & 0.621/0.210 & 0.745/0.260 & 0.698/0.241
& 0.737/0.306 & 0.018 \\
ICP-force
& 1.038/0.290 & 0.516/0.229 & 0.602/0.226 & 0.702/0.285 & 0.671/0.263
& 0.706/0.299 & 0.024 \\
Covariance T2G~\cite{bimbo2016covariance}
& 0.530/0.319 & 0.857/0.821 & 0.646/0.402 & 0.912/0.613 & 0.748/0.521
& 0.739/0.635 & 0.092 \\
Contact-SDF PF
& 0.487/0.238 & 0.512/0.171 & 0.452/0.162 & 0.562/0.218 & 0.496/0.189
& 0.502/0.209 & 0.118 \\
Single-frame Physics PF ($W=1$)
& 0.590/0.238 & 0.545/0.210 & 0.522/0.196 & 0.624/0.250 & 0.557/0.223
& 0.568/0.214 & 0.156 \\
Dikhale-TactileOnly$^\ast$~\cite{dikhaleVisuoTactile6DPose2022a}
& 0.895/0.420 & 0.730/0.350 & 0.585/0.260 & 0.812/0.410 & 0.768/0.360
& 0.758/0.381 & 0.047 \\
Ours ($W=5$)
& \textbf{0.342/0.200} & \textbf{0.407/0.084} & \textbf{0.311/0.095}
& \textbf{0.386/0.122} & \textbf{0.359/0.118}
& \textbf{0.361/0.124} & 0.480 \\
\bottomrule
\end{tabular}
\vspace{-10pt}
\end{table*}

% =========================
% End of Section III
% =========================
%%%%%%%%%%%%%%%%%%%%%%%%%%%%%%%%%%%%%%%%%%%%%%%%%%%%%%%%%%%%%%%%%%%%%%%%%%%%%%%%
% =========================
% Section IV: Experiments
% =========================
\section{Experiments}
\label{sec:experiments}

\subsection{Experimental Setup}
\label{subsec:experimental_setup}

\begin{figure}[!h]
\centering
\includegraphics[width=0.3\textwidth]{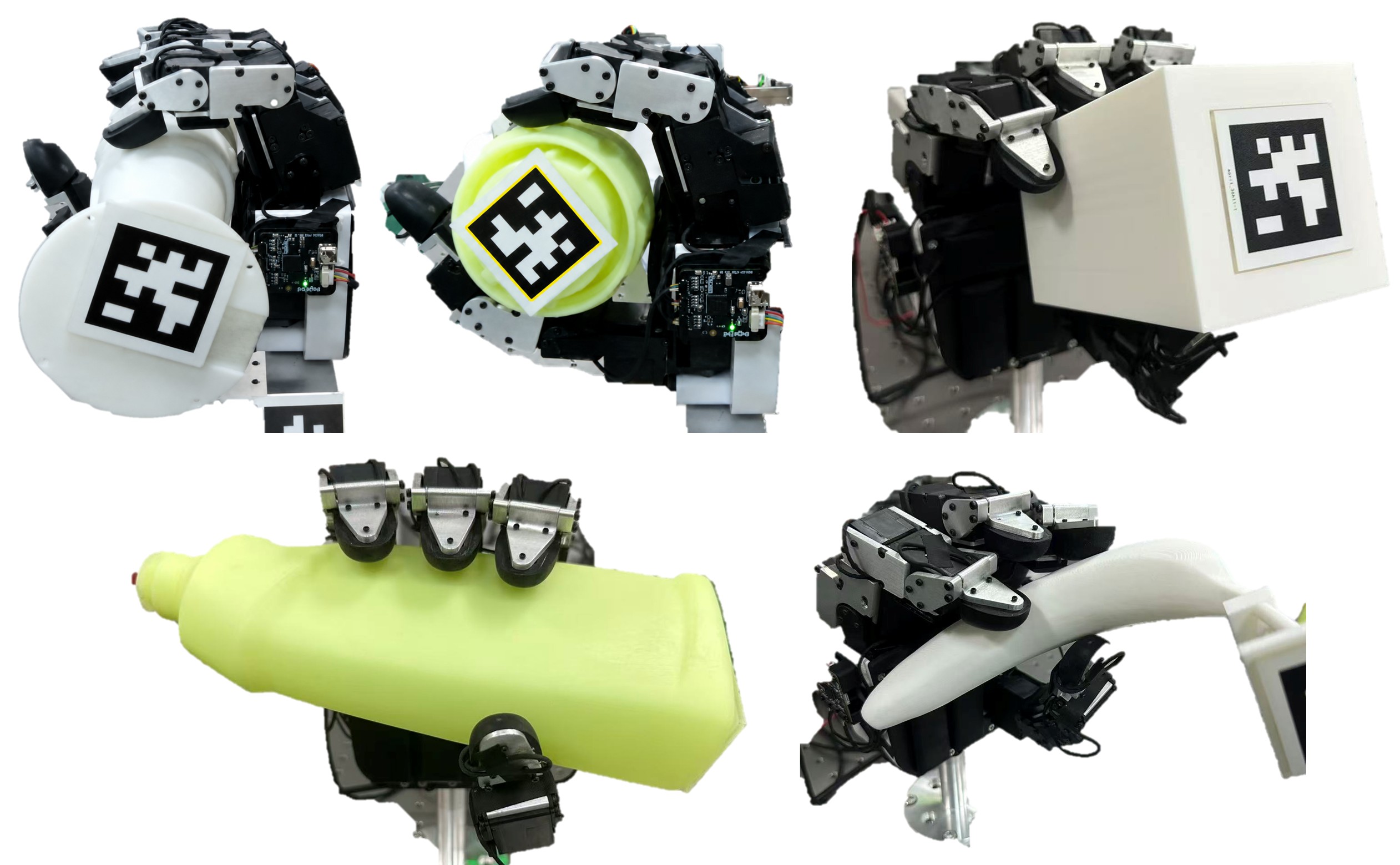}
\caption{Representative configurations of the screwdriver, connector, cuboid, flat bottle, and banana on the tactilely instrumented Allegro Hand V5.}
\label{fig:hand}
\end{figure}

Experiments are conducted on an Allegro Hand V5 equipped with 16 PaXini tactile modules distributed over the fingers and palm. Each taxel provides an estimated 3D contact force, and taxel positions are transformed into a common hand/world frame using calibrated forward kinematics. Each manipulated object is represented by a watertight mesh, from which nearest surface points, normals, and signed distances are queried online. Fig.~\ref{fig:hand} shows representative grasp configurations of all evaluated objects on the tactilely instrumented hand.

We evaluate five rigid objects: an electric screwdriver, electrical connector, cuboid, flat bottle, and banana. The flat bottle and banana are from YCB~\cite{calliBenchmarkingManipulationResearch2015}; the other three meshes are reconstructed from the corresponding physical objects. Short quasi-static teleoperated sequences vary placement, orientation, and grasp. Elongated objects mainly use palm-supported grasps; compact objects also use pinch grasps when applicable. Rigidly attached AprilTags provide ground truth only for evaluation.

\begin{table}[!t]
\centering
\caption{Summary of the evaluated object set and tactile sequences.}
\label{tab:object_sequences}
\begin{tabular}{lcc}
\toprule
Object & $s_o$ / Grasp type & Trials / frames per trial \\
\midrule
Screwdriver & 33.7, only grasp & 200 / 240 \\
Connector & 12.8, grasp \& pinch & 200 / 240 \\
Cuboid & 18.0, grasp \& pinch & 200 / 240 \\
Flat bottle & 26.8, only grasp & 200 / 240 \\
Banana & 20.4, only pinch & 200 / 240 \\
\bottomrule
\end{tabular}
\end{table}

We use normalized ADD-S as the primary metric. For each object, the raw ADD-S is divided by the object scale \(s_o=\|\mathbf{b}^{o}_{\max}-\mathbf{b}^{o}_{\min}\|_2\), where \(\mathbf{b}^{o}_{\max}\) and \(\mathbf{b}^{o}_{\min}\) are the opposite corners of the axis-aligned bounding box computed from the object mesh. We report the median and interquartile range (IQR) over repeated trials to characterize accuracy and robustness.

Throughout Tables~\ref{tab:baseline_expanded}--\ref{tab:ablation_relocalization}, one trial is one refinement episode initialized at a tactile sequence start. Each object has 200 episodes of 240 recorded frames; at most \(N_{\mathrm{upd}}^{\max}=120\) scheduled filter updates are processed with identical budgets across variants. The pose is treated as constant within an update/window, while slow changes may occur between quasi-static configurations. An episode fails if it exhausts the budget before the ``Conv. steps'' stopping rule, or if its final dominant-mode estimate has \(\bar d_{\mathrm{pen}}>0.03\,\mathrm{m}\), or still has \(r_{\mathrm{ESS}}<0.005\) and \(\ell_{\mathrm{best}}<-10^6\). A relocalization trigger starts recovery; only the final episode state determines failure. ADD-S uses every episode's final estimate, including failures; ``Conv. steps'' averages successful episodes only.

Unless otherwise stated, the full model uses \(N=500\) particles, window length \(W=5\), and the full likelihood in Eq.~\eqref{eq:full_likelihood}. On the i9-13950HX/RTX4060 laptop, one update takes \(2.12\,\mathrm{s}\) on CPU and \(0.48\,\mathrm{s}\) on GPU; therefore, the current implementation is positioned as low-rate quasi-static pose refinement rather than high-speed dynamic tracking.

\subsection{Main Comparisons with Baselines}

We compare the proposed method with tactile-only baselines under the same sensing assumptions: integrated whole-hand tactile arrays, known object meshes, and no external visual input during inference. The compared methods include point-to-point ICP, ICP-force, Covariance T2G~\cite{bimbo2016covariance}, Contact-SDF PF, Single-frame Physics PF, and Dikhale-TactileOnly$^\ast$~\cite{dikhaleVisuoTactile6DPose2022a}.

ICP aligns active taxel centers to nearest object-surface points using point-to-point nearest-neighbor matching. We use a maximum of 150 ICP iterations and terminate early when the relative change of the mean squared nearest-neighbor residual falls below $10^{-5}$. ICP-force uses the same ICP settings, but first displaces each active taxel along the opposite normalized force direction,
\begin{equation}
\mathbf{p}'_k
=
\mathbf{p}_k
-
\alpha_{\rm ICP}\|\mathbf{f}_k\|_2 \hat{\mathbf{f}}_k,
\quad
\hat{\mathbf{f}}_k
=
\frac{\mathbf{f}_k}{\|\mathbf{f}_k\|_2+\epsilon}.
\label{eq:icp_force}
\end{equation}

The coefficient $\alpha_{\rm ICP}$ is selected once on a validation subset and fixed for all objects. Covariance T2G uses the same active taxels and object meshes as our method, with fixed coarse-to-fine search, orientation discretization, and local covariance-window settings. Contact-SDF PF uses the same particle representation as our method but retains only the active-contact signed-distance term. Single-frame Physics PF keeps the full physics-informed likelihood but disables temporal fusion by setting $W=1$. Dikhale-TactileOnly$^\ast$ removes all RGB-D inputs from the original visuo-tactile setting and uses only the active-taxel contact point cloud as input.

All methods use the same tactile preprocessing, taxel selection, coordinate-frame conversion, synchronized ground-truth poses, object-wise trial splits, and normalized ADD-S evaluation. Methods requiring initialization use the previous-frame estimate, with identity initialization at the first frame. No baseline parameters are tuned separately for individual test sequences.

Table~\ref{tab:baseline_expanded} reports the expanded five-object results. The proposed method achieves the best average median/IQR, \(0.361/0.124\). Gains over ICP, ICP-force, and Covariance T2G show that sparse active-contact geometry alone is insufficient; the two internal ablations isolate the benefits of physics-informed feasibility and temporal fusion.

The proposed method is slower because it evaluates a multi-term likelihood over particles and a sliding window. The dominant cost is repeated nearest-surface, normal, and SDF queries; excluding optional guidance, analytical complexity is approximately \(\mathcal{O}(NWKQ_{\mathrm{mesh}})\), where \(Q_{\mathrm{mesh}}\) is average mesh-query cost. Comparators were selected for sensing compatibility and reproducibility, not equal runtime. More iterations may improve their convergence but cannot add unavailable force/temporal evidence; because we did not run a matched-compute sweep, we make no claim that their accuracy cannot improve. The internal ablations require 0.118--0.156~s (25--33\% of full-model time) and offer lower-cost operating points; the full \(N=500,W=5\) setting is accuracy-oriented.

\subsection{Ablation Study}

We conduct grouped ablations and proposal-scale sensitivity tests on the screwdriver and connector under the protocol in Sec.~\ref{subsec:experimental_setup}. Table~\ref{tab:ablation_sensitivity} reports statistics pooled over their 400 episodes (not an average of object medians); its full-model row is also the \(1.0\times\) nominal proposal setting. Each group isolates one claimed component or auxiliary mechanism.

\begin{table}[!t]
\centering
\caption{Ablation and proposal-scale sensitivity on screwdriver and connector sequences.}
\label{tab:ablation_sensitivity}
\scriptsize
\setlength{\tabcolsep}{2.6pt}
\begin{tabular}{@{}lccc@{}}
\toprule
Setting & ADD-S Med. / IQR & Conv. steps & Fail. rate \\
\midrule
Full model / $1.0\times$ nominal & \textbf{0.375 / 0.176} & \textbf{12.4} & \textbf{7.6\%} \\
\midrule
\multicolumn{4}{l}{\textit{Component ablations}} \\
Contact-SDF PF & 0.461 / 0.235 & 21.3 & 14.2\% \\
Single-frame Physics PF ($W=1$) & 0.568 / 0.224 & 19.8 & 15.6\% \\
Without potential guidance & 0.421 / 0.206 & 45.1 & 16.8\% \\
Without mode preservation & 0.426 / 0.218 & -- & 29.0\% \\
\midrule
\multicolumn{4}{l}{\textit{Proposal-scale sensitivity}} \\
Isotropic proposal & 0.402 / 0.193 & 16.9 & 9.7\% \\
$0.5\times$ nominal & 0.391 / 0.185 & 21.6 & 8.1\% \\
$2.0\times$ nominal & 0.414 / 0.207 & 14.8 & 12.1\% \\
\bottomrule
\end{tabular}
\end{table}

\textit{Physics-informed feasibility.}
Contact-SDF PF tests geometric contact consistency without the additional physics terms. Table~\ref{tab:ablation_sensitivity} shows that these terms reduce median ADD-S from 0.461 to 0.375 (18.7\%) and IQR from 0.235 to 0.176 (25.1\%) by suppressing implausible poses.

\begin{figure}[!t]
    \centering
    \includegraphics[width=0.34\textwidth]{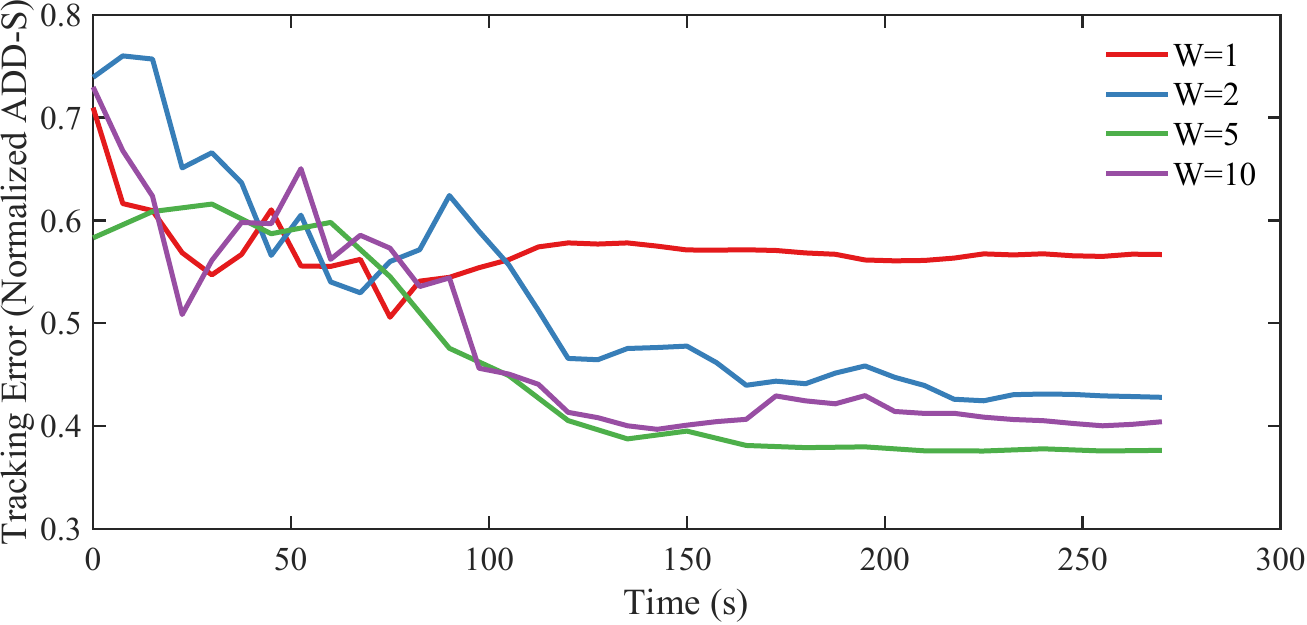}
    \vspace{-10pt}
    \caption{Effect of sliding-window length. Mean normalized ADD-S for $W=1,2,5,10$ over 400 trajectories (200 per object) from the screwdriver and connector sets.}
    \label{fig:ablation_window_curve}
\end{figure}

\textit{Sliding-window fusion.}
Removing the temporal window by setting $W=1$ increases the median/IQR normalized ADD-S from 0.375/0.176 to 0.568/0.224. Fig.~\ref{fig:ablation_window_curve} further shows that larger windows produce smoother and generally lower-error trajectories than the single-frame update, while $W=5$ provides a practical balance between accuracy and computation.

\textit{Potential-field guidance.}
We ablate the potential-field-guided propagation by disabling Eq.~\eqref{eq:potential_drift_update} while keeping all other settings unchanged. As shown in Table~\ref{tab:ablation_sensitivity}, removing this guidance increases the average convergence length from 12.4 to 45.1 filter updates, indicating that the guidance mainly improves sample efficiency by steering particles away from implausible hand--object penetration regions.

\textit{Symmetry-aware mode preservation.}
For the near-axisymmetric objects, disabling mode preservation worsens ADD-S from 0.375/0.176 to 0.426/0.218, translation from 7.76 to 8.66~mm, and axis-aware rotation from $6.55^\circ$ to $12.83^\circ$, while success falls from 92\% to 71\%. Thus, retaining competing yaw modes avoids premature collapse under sparse local contacts.

\textit{Proposal design and scale.}
The $0.5\times$ and $2.0\times$ settings multiply all initial/final translational and rotational standard deviations in Eqs.~\eqref{eq:trans_aniso}--\eqref{eq:rot_aniso} by the stated factor; all other settings and budgets remain fixed. Performance is stable near nominal, indicating that anisotropy and scale chiefly affect convergence speed and dispersion rather than likelihood validity.

\textit{Relocalization.}
Table~\ref{tab:ablation_relocalization} evaluates degeneracy-triggered relocalization. The trigger rate reports the percentage of episodes with at least one relocalization; a triggered episode is counted as a failure only if its final estimate still meets the failure criteria in Sec.~\ref{subsec:experimental_setup}. Relocalization is infrequent in nominal sequences (2.4\%), so it is not the main source of average accuracy gain. It nevertheless reduces the failure rate from 6.1\% to 2.6\% in nominal sequences and from 28.7\% to 8.2\% under perturbed initialization, supporting its role as a recovery mechanism for particle collapse and severe initialization errors.

\begin{table}[!t]
\centering
\caption{Relocalization ablation; perturbed initialization adds random 30-mm translation and $30^\circ$ rotation offsets.}
\label{tab:ablation_relocalization}
\scriptsize
\begin{tabular}{llccc}
\toprule
Setting & Variant & med. ADD-S / IQR & Fail. rate & Trig. rate \\
% & & med. / IQR & rate & rate \\
\midrule
Nominal & w/o reloc. & 0.386 / 0.153 & 6.1\% & -- \\
Nominal & Ours & \textbf{0.361 / 0.124} & \textbf{2.6\%} & 2.4\% \\
Pert. init. & w/o reloc. & 0.684 / 0.336 & 28.7\% & -- \\
Pert. init. & Ours & \textbf{0.418 / 0.179} & \textbf{8.2\%} & 12.1\% \\
\bottomrule
\end{tabular}
\end{table}

% =========================
% End of Section IV
% =========================
%%%%%%%%%%%%%%%%%%%%%%%%%%%%%%%%%%%%%%%%%%%%%%%%%%%%%%%%%%%%%%%%%%%%%%%%%%%%%%%%
% =========================
% Section V: Discussion and Conclusions
% =========================
\section{Discussion and Conclusions}
\label{sec:discussion_conclusion}

We presented a physics-informed sliding-window particle filter for tactile-only in-hand 6-DoF pose refinement from whole-hand 3D force taxels. On static and short quasi-static sequences, the complete formulation achieves lower normalized ADD-S than the evaluated tactile-only baselines. The ablations show complementary roles: physical feasibility suppresses implausible hypotheses; temporal fusion reduces intermittent-contact ambiguity; penetration-aware guidance improves sample efficiency; and symmetry-aware resampling prevents premature yaw-mode collapse. Relocalization remains an auxiliary recovery mechanism rather than the main source of nominal accuracy.

The intended operating point is low-rate refinement and belief maintenance, not high-speed dynamic tracking. At 0.480~s/update, the full $N=500,W=5$ setting is accuracy-oriented; Contact-SDF PF and Single-frame Physics PF require 0.118--0.156~s/update. Further savings are possible through fewer particles, a shorter window, zero-force taxel subsampling, less frequent guidance, or disabling auxiliary guards, at reduced robustness. The $\mathcal{O}(NWKQ_{\mathrm{mesh}})$ complexity is analytical: no measured $N/W/K$ curve or matched-compute baseline sweep was conducted, so additional computation may improve comparator convergence.

The main limitations are a known rigid watertight mesh, calibrated taxel poses/force directions, and approximately constant pose within each update/window. Mesh error, force bias, unmodeled compliance, or appreciable inertial effects can distort the likelihood. Symmetry treatment is restricted to a single or dominant rotational axis, and repeated SDF/nearest-surface queries remain the bottleneck. The five objects span distinct geometries but do not establish category-level generalization; moreover, ADD-S does not directly measure downstream manipulation success. Future work will study broader shapes and symmetries, task-level metrics, faster motion, uncertain shape, online physical-parameter calibration, measured accuracy--runtime scaling, and visuo-tactile integration. Public synchronized sequences and reference implementations would improve reproducibility under common budgets.

% =========================
% End of Section V
% =========================
%%%%%%%%%%%%%%%%%%%%%%%%%%%%%%%%%%%%%%%%%%%%%%%%%%%%%%%%%%%%%%%%%%%%%%%%%%%%%%%%
%%%%%%%%%%%%%%%%%%%%%%%%%%%%%%%%%%%%%%%%%%%%%%%%%%%%%%%%%%%%%%%%%%%%%%%%%%%%%%%%
% \begin{thebibliography}{99}

\bibliographystyle{ref/IEEEtran}
\bibliography{ref/ieee2026}

@inproceedings{alvarezTactilebasedInhandObject2018a,
  title = {Tactile-Based in-Hand Object Pose Estimation},
  booktitle = {{{ROBOT}} 2017: {{Third Iberian Robotics Conference}}},
  author = {{\'A}lvarez, David and Roa, M{\'a}ximo A. and Moreno, Luis},
  editor = {Ollero, Anibal and Sanfeliu, Alberto and Montano, Luis and Lau, Nuno and Cardeira, Carlos},
  year = 2018,
  pages = {716--728},
  publisher = {Springer International Publishing},
  address = {Cham},
  doi = {10.1007/978-3-319-70836-2_59},
  isbn = {978-3-319-70836-2}
}

@article{barfootAssociatingUncertaintyThreedimensional2014,
  title = {Associating Uncertainty with Three-Dimensional Poses for Use in Estimation Problems},
  author = {Barfoot, Timothy D. and Furgale, Paul T.},
  year = 2014,
  month = jun,
  journal = {IEEE Transactions on Robotics},
  volume = {30},
  number = {3},
  pages = {679--693},
  doi = {10.1109/TRO.2014.2298059},
}

@article{barrauInvariantExtendedKalman2017,
  title = {The Invariant Extended Kalman Filter as a Stable Observer},
  author = {Barrau, Axel and Bonnabel, Silv{\`e}re},
  year = 2017,
  month = apr,
  journal = {IEEE Transactions on Automatic Control},
  volume = {62},
  number = {4},
  pages = {1797--1812},
  doi = {10.1109/TAC.2016.2594085},
}

@article{barrauInvariantKalmanFiltering2018,
  author  = {Barrau, Axel and Bonnabel, Silv{\`e}re},
  title   = {Invariant Kalman Filtering},
  journal = {Annual Review of Control, Robotics, and Autonomous Systems},
  year    = {2018},
  volume  = {1},
  pages   = {237--257},
  month   = may,
  doi     = {10.1146/annurev-control-060117-105010}
}

@article{bauzaTac2PoseTactileObject2023a,
  title = {{{Tac2Pose}}: {{Tactile}} Object Pose Estimation from the First Touch},
  author = {Bauza, Maria and Bronars, Antonia and Rodriguez, Alberto},
  year = 2023,
  month = nov,
  journal = {The International Journal of Robotics Research},
  volume = {42},
  number = {13},
  pages = {1185--1209},
  publisher = {SAGE Publications Ltd STM},
  doi = {10.1177/02783649231196925},
}

@inproceedings{caddeoCollisionawareInhand6D2023b,
  title = {Collision-Aware in-Hand {{6D}} Object Pose Estimation Using Multiple Vision-Based Tactile Sensors},
  author = {Caddeo, Gabriele M. and Piga, Nicola A. and Bottarel, Fabrizio and Natale, Lorenzo},
  year = 2023,
  month = may,
  booktitle = {2023 IEEE International Conference on Robotics and Automation (ICRA)},
  pages = {719--725},
  publisher = {IEEE},
  address = {London, United Kingdom},
  doi = {10.1109/ICRA48891.2023.10160359},
  isbn = {9798350323658}
}

@article{calliBenchmarkingManipulationResearch2015,
  title = {Benchmarking in Manipulation Research: Using the Yale-{{CMU-berkeley}} Object and Model Set},
  author = {Calli, Berk and Walsman, Aaron and Singh, Arjun and Srinivasa, Siddhartha and Abbeel, Pieter and Dollar, Aaron M.},
  year = 2015,
  month = sep,
  journal = {IEEE Robotics and Automation Magazine},
  volume = {22},
  number = {3},
  pages = {36--52},
  doi = {10.1109/MRA.2015.2448951},
}

@inproceedings{chalonOnlineInhandObject2013a,
  title = {Online In-Hand Object Localization},
  booktitle = {2013 {{IEEE}}/{{RSJ International Conference}} on {{Intelligent Robots}} and {{Systems}}},
  author = {Chalon, Maxime and Reinecke, Jens and Pfanne, Martin},
  year = 2013,
  month = nov,
  pages = {2977--2984},
  doi = {10.1109/IROS.2013.6696778},
}

@article{bimbo2016covariance,
  title={In-Hand Object Pose Estimation Using Covariance-Based Tactile To Geometry Matching},
  author={Bimbo, Joao and Luo, Shan and Althoefer, Kaspar and Liu, Hongbin},
  journal={IEEE Robotics and Automation Letters},
  volume={1},
  number={1},
  pages={570--577},
  year={2016}
}

@inproceedings{dellaertMonteCarloLocalization1999,
  title = {Monte Carlo Localization for Mobile Robots},
  booktitle = {Proceedings 1999 {{IEEE International Conference}} on {{Robotics}} and {{Automation}} ({{Cat}}. {{No}}.{{99CH36288C}})},
  author = {Dellaert, F. and Fox, D. and Burgard, W. and Thrun, S.},
  year = 1999,
  month = may,
  volume = {2},
  pages = {1322-1328 vol.2},
  doi = {10.1109/ROBOT.1999.772544},
}

@article{dikhaleVisuoTactile6DPose2022a,
  title = {{{VisuoTactile 6D}} Pose Estimation of an {{In-hand}} Object Using Vision and Tactile Sensor Data},
  author = {Dikhale, Snehal and Patel, Karankumar and Dhingra, Daksh and Naramura, Itoshi and Hayashi, Akinobu and Iba, Soshi and Jamali, Nawid},
  year = 2022,
  month = apr,
  journal = {IEEE Robotics and Automation Letters},
  volume = {7},
  number = {2},
  pages = {2148--2155},
  doi = {10.1109/LRA.2022.3143289},
}

@inproceedings{donlonGelSlimHighresolutionCompact2018a,
  title = {{{GelSlim}}: A High-Resolution, Compact, Robust, and Calibrated Tactile-Sensing Finger},
  booktitle = {2018 {{IEEE}}/{{RSJ International Conference}} on {{Intelligent Robots}} and {{Systems}} ({{IROS}})},
  author = {Donlon, Elliott and Dong, Siyuan and Liu, Melody and Li, Jianhua and Adelson, Edward and Rodriguez, Alberto},
  year = 2018,
  month = oct,
  pages = {1927--1934},
  doi = {10.1109/IROS.2018.8593661},
}

@article{hartleyContactaidedInvariantExtended2020,
  title = {Contact-Aided Invariant Extended Kalman Filtering for Robot State Estimation},
  author = {Hartley, Ross and Ghaffari, Maani and Eustice, Ryan M and Grizzle, Jessy W},
  year = 2020,
  month = mar,
  journal = {The International Journal of Robotics Research},
  volume = {39},
  number = {4},
  pages = {402--430},
  publisher = {SAGE Publications Ltd STM},
  doi = {10.1177/0278364919894385},
}

@article{kaessISAM2IncrementalSmoothing2012,
  title = {{{iSAM2}}: {{Incremental}} Smoothing and Mapping Using the Bayes Tree},
  author = {Kaess, Michael and Johannsson, Hordur and Roberts, Richard and Ila, Viorela and Leonard, John J and Dellaert, Frank},
  year = 2012,
  month = feb,
  journal = {The International Journal of Robotics Research},
  volume = {31},
  number = {2},
  pages = {216--235},
  doi = {10.1177/0278364911430419},
}

@inproceedings{kimSimultaneousTactileEstimation2023,
  title = {Simultaneous Tactile Estimation and Control of Extrinsic Contact},
  booktitle = {2023 {{IEEE International Conference}} on {{Robotics}} and {{Automation}} ({{ICRA}})},
  author = {Kim, Sangwoon and Jha, Devesh K. and Romeres, Diego and Patre, Parag and Rodriguez, Alberto},
  year = 2023,
  month = may,
  pages = {12563--12569},
  doi = {10.1109/ICRA48891.2023.10161158},
}

@inproceedings{kovalManifoldParticleFilter2017,
  title = {The Manifold Particle Filter for State Estimation on High-Dimensional Implicit Manifolds},
  booktitle = {2017 {{IEEE International Conference}} on {{Robotics}} and {{Automation}} ({{ICRA}})},
  author = {Koval, Michael C. and Klingensmith, Matthew and Srinivasa, Siddhartha S. and Pollard, Nancy S. and Kaess, Michael},
  year = 2017,
  month = may,
  pages = {4673--4680},
  doi = {10.1109/ICRA.2017.7989543},
}

@inproceedings{kovalPoseEstimationContact2013a,
  title = {Pose Estimation for Contact Manipulation with Manifold Particle Filters},
  booktitle = {2013 {{IEEE}}/{{RSJ International Conference}} on {{Intelligent Robots}} and {{Systems}}},
  author = {Koval, Michael C. and Dogar, Mehmet R. and Pollard, Nancy S. and Srinivasa, Siddhartha S.},
  year = 2013,
  month = nov,
  pages = {4541--4548},
  doi = {10.1109/IROS.2013.6697009},
}

@article{lambetaDIGITNovelDesign2020a,
  title = {{{DIGIT}}: A Novel Design for a Low-Cost Compact High-Resolution Tactile Sensor with Application to {{In-hand}} Manipulation},
  author = {Lambeta, Mike and Chou, Po-Wei and Tian, Stephen and Yang, Brian and Maloon, Benjamin and Most, Victoria Rose and Stroud, Dave and Santos, Raymond and Byagowi, Ahmad and Kammerer, Gregg and Jayaraman, Dinesh and Calandra, Roberto},
  year = 2020,
  month = jul,
  journal = {IEEE Robotics and Automation Letters},
  volume = {5},
  number = {3},
  pages = {3838--3845},
  doi = {10.1109/LRA.2020.2977257},
}

@inproceedings{leeViTaSCOPEVisuotactileImplicit2025b,
  title = {{{ViTaSCOPE}}: {{Visuo-tactile}} Implicit Representation for in-Hand Pose and Extrinsic Contact Estimation},
  booktitle = {3rd {{RSS Workshop}} on {{Dexterous Manipulation}}: {{Learning}} and {{Control}} with {{Diverse Data}}},
  author = {Lee, Jayjun and Fazeli, Nima},
  year = 2025,
  month = jun,
}

@article{liuEnhancingGeneralizable6D2024a,
  title = {Enhancing Generalizable {{6D}} Pose Tracking of an {{In-hand}} Object with Tactile Sensing},
  author = {Liu, Yun and Xu, Xiaomeng and Chen, Weihang and Yuan, Haocheng and Wang, He and Xu, Jing and Chen, Rui and Yi, Li},
  year = 2024,
  month = feb,
  journal = {IEEE Robotics and Automation Letters},
  volume = {9},
  number = {2},
  pages = {1106--1113},
  doi = {10.1109/LRA.2023.3337690},
}

@article{liViHOPEVisuotactileInhand2023,
  title = {{{ViHOPE}}: Visuotactile {{In-hand}} Object {{6D}} Pose Estimation with Shape Completion},
  author = {Li, Hongyu and Dikhale, Snehal and Iba, Soshi and Jamali, Nawid},
  year = 2023,
  month = nov,
  journal = {IEEE Robotics and Automation Letters},
  volume = {8},
  number = {11},
  pages = {6963--6970},
  doi = {10.1109/LRA.2023.3313941},
}

@article{luoRoboticTactilePerception2017a,
  title = {Robotic Tactile Perception of Object Properties: A Review},
  author = {Luo, Shan and Bimbo, Joao and Dahiya, Ravinder and Liu, Hongbin},
  year = 2017,
  month = dec,
  journal = {Mechatronics},
  volume = {48},
  pages = {54--67},
  doi = {10.1016/j.mechatronics.2017.11.002},
}

@article{petrovskayaGlobalLocalizationObjects2011a,
  title = {Global Localization of Objects via Touch},
  author = {Petrovskaya, Anna and Khatib, Oussama},
  year = 2011,
  month = jun,
  journal = {IEEE Transactions on Robotics},
  volume = {27},
  number = {3},
  pages = {569--585},
  doi = {10.1109/TRO.2011.2138450},
}

@inproceedings{pfanneEKFbasedInhandObject2017a,
  title = {{{EKF-based}} in-Hand Object Localization from Joint Position and Torque Measurements},
  booktitle = {2017 {{IEEE}}/{{RSJ International Conference}} on {{Intelligent Robots}} and {{Systems}} ({{IROS}})},
  author = {Pfanne, Martin and Chalon, Maxime},
  year = 2017,
  month = sep,
  pages = {2464--2470},
  doi = {10.1109/IROS.2017.8206063},
}

@inproceedings{siposSimultaneousContactLocation2022a,
  title = {Simultaneous Contact Location and Object Pose Estimation Using Proprioception and Tactile Feedback},
  booktitle = {2022 {{IEEE}}/{{RSJ International Conference}} on {{Intelligent Robots}} and {{Systems}} ({{IROS}})},
  author = {Sipos, Andrea and Fazeli, Nima},
  year = 2022,
  month = oct,
  pages = {3233--3240},
  doi = {10.1109/IROS47612.2022.9981762},
}

@article{vandermerweSimultaneousExtrinsicContact2026,
  title = {Simultaneous Extrinsic Contact and {{In-hand}} Pose Estimation via Distributed Tactile Sensing},
  author = {{Van der Merwe}, Mark and Ota, Kei and Berenson, Dmitry and Fazeli, Nima and Jha, Devesh K.},
  year = 2026,
  month = mar,
  journal = {IEEE Robotics and Automation Letters},
  volume = {11},
  number = {3},
  pages = {2394--2401},
  doi = {10.1109/LRA.2026.3653324},
}

@inproceedings{villalongaTactileObjectPose2021b,
  title = {Tactile Object Pose Estimation from the First Touch with Geometric Contact Rendering},
  booktitle = {Proceedings of the 2020 {{Conference}} on {{Robot Learning}}},
  author = {Villalonga, Maria Bauza and Rodriguez, Alberto and Lim, Bryan and Valls, Eric and Sechopoulos, Theo},
  year = 2021,
  month = oct,
  pages = {1015--1029},
  publisher = {PMLR},
}

@inproceedings{wangDenseFusion6DObject2019a,
  title = {{{DenseFusion}}: {{6D}} Object Pose Estimation by Iterative Dense Fusion},
  author = {Wang, Chen and Xu, Danfei and Zhu, Yuke and {Mart{\'i}n-Mart{\'i}n}, Roberto and Lu, Cewu and {Fei-Fei}, Li and Savarese, Silvio},
  year = 2019,
  month = jun,
  booktitle = {2019 IEEE/CVF Conference on Computer Vision and Pattern Recognition (CVPR)},
  pages = {3338--3347},
  publisher = {IEEE},
  address = {Long Beach, CA, USA},
  doi = {10.1109/CVPR.2019.00346},
  isbn = {9781728132938}
}

@article{ward-cherrierTacTipFamilySoft2018,
  title = {The {{TacTip}} Family: Soft Optical Tactile Sensors with {{3D-printed}} Biomimetic Morphologies},
  author = {{Ward-Cherrier}, Benjamin and Pestell, Nicholas and Cramphorn, Luke and Winstone, Benjamin and Giannaccini, Maria Elena and Rossiter, Jonathan and Lepora, Nathan F.},
  year = 2018,
  month = apr,
  journal = {Soft Robotics},
  volume = {5},
  number = {2},
  pages = {216--227},
  doi = {10.1089/soro.2017.0052},
  pmcid = {PMC5905869},
  pmid = {29297773}
}

@inproceedings{xiangPoseCNNConvolutionalNeural2018b,
  title = {{{PoseCNN}}: {{A}} Convolutional Neural Network for {{6D}} Object Pose Estimation in Cluttered Scenes},
  author = {Xiang, Yu and Schmidt, Tanner and Narayanan, Venkatraman and Fox, Dieter},
  year = 2018,
  month = jun,
  booktitle = {Robotics: Science and Systems XIV},
  publisher = {{Robotics: Science and Systems Foundation}},
  doi = {10.15607/RSS.2018.XIV.019},
  isbn = {9780992374747}
}

@article{yousefTactileSensingDexterous2011,
  title = {Tactile Sensing for Dexterous In-Hand Manipulation in Robotics---a Review},
  author = {Yousef, Hanna and Boukallel, Mehdi and Althoefer, Kaspar},
  year = 2011,
  month = jun,
  journal = {Sensors and Actuators A: Physical},
  series = {Solid-{{State Sensors}}, {{Actuators}} and {{Microsystems Workshop}}},
  volume = {167},
  number = {2},
  pages = {171--187},
  doi = {10.1016/j.sna.2011.02.038},
}

@article{yuanGelSightHighresolutionRobot2017a,
  title = {{{GelSight}}: High-Resolution Robot Tactile Sensors for Estimating Geometry and Force},
  author = {Yuan, Wenzhen and Dong, Siyuan and Adelson, Edward H.},
  year = 2017,
  month = dec,
  journal = {Sensors},
  volume = {17},
  number = {12},
  pages = {2762},
  publisher = {Multidisciplinary Digital Publishing Institute},
  doi = {10.3390/s17122762},
}
% \end{thebibliography}
\end{document}